%% file: main.tex
\pdfoutput=1

\documentclass[11pt]{article}

\usepackage[final]{acl}

\usepackage{times}
\usepackage{latexsym}

\input{sources/math_commands.tex}

\input{preamble}

\usepackage[T1]{fontenc}

\usepackage[utf8]{inputenc}

\usepackage{microtype}

\usepackage{inconsolata}

\usepackage{graphicx}
\usepackage{url}            
\usepackage{booktabs}       
\usepackage{amsfonts}       
\usepackage{nicefrac}       
\usepackage{microtype}      
\usepackage{xcolor}         

\usepackage{multirow}
\usepackage{adjustbox}
\usepackage{graphicx}
\usepackage{array}
\usepackage{amsmath}

\usepackage{tabularx}
\usepackage{xspace}
\usepackage{xcolor}         
\definecolor{citecolor}{rgb}{0.21,0.49,0.74}
\definecolor{linkcolor}{rgb}{0.8,0.16,0.16}

\makeatletter
\DeclareRobustCommand\onedot{\futurelet\@let@token\@onedot}
\def\@onedot{\ifx\@let@token.\else.\null\fi\xspace}

\def\eg{\emph{e.g}\onedot} 
\def\ie{\emph{i.e}\onedot} 
 
\def\etc{\emph{etc}\onedot} 
\def\wrt{w.r.t\onedot} 

\makeatother

\usepackage[capitalize]{cleveref}
    \crefname{section}{Sec.}{Secs.}
    \Crefname{section}{Section}{Sections}
    \Crefname{table}{Table}{Tables}
    \crefname{table}{Table}{Tables} 

\makeatletter
\def\NAT@spacechar{~}
\makeatother

\usepackage{enumitem}
\setlist[itemize]{leftmargin=5mm, itemsep=0.0mm}

\usepackage[nottoc]{tocbibind}
\usepackage{minitoc}
\usepackage{bbm}

\usepackage{pifont}
\usepackage{float}
\newcommand{\cmark}{\ding{51}}%
\newcommand{\xmark}{\ding{55}}%
\usepackage{blindtext}
\usepackage{multicol}
\usepackage{caption}
\usepackage{subcaption}
\usepackage{float}

\usepackage[nottoc]{tocbibind}
\usepackage{minitoc}
\usepackage{bbm}

\title{In-Depth and In-Breadth: Pre-training Multimodal Language Models \\ Customized for Comprehensive Chart Understanding}

\author{
 \textbf{Wan-Cyuan Fan\textsuperscript{1,3}}$^*$,
 \textbf{Yen-Chun Chen\textsuperscript{2}},
 \textbf{Mengchen Liu\textsuperscript{2}},
 \textbf{Alexander Jacobson\textsuperscript{1}},
\\
 \textbf{Lu Yuan\textsuperscript{2}},
 \textbf{Leonid Sigal\textsuperscript{1,3,4}}
\\
\\
 \textsuperscript{1}UBC,
 \textsuperscript{2}Microsoft,
 \textsuperscript{3}Vector Institute for AI,
 \textsuperscript{4}CIFAR AI Chair
\\
 \small{
   \textbf{Correspondence:} \href{mailto:email@domain}{wancyuan@cs.ubc.ca}
 }
}
\begin{document}
\maketitle
\begin{abstract}
\input{sections/0_abstract}
\end{abstract}

\input{sections/1_introduction}

\input{sections/2_related_work}

\input{sections/3_method}

\input{sections/4_exp}
\input{sections/5_conclusion}

\bibliography{sources/iclr2025_conference}



\end{document}

%% file: sources/math_commands.tex
\usepackage{amsmath,amsfonts,bm}

\def\eqref#1{equation~\ref{#1}}

\def\1{\bm{1}}

\DeclareMathAlphabet{\mathsfit}{\encodingdefault}{\sfdefault}{m}{sl}
\SetMathAlphabet{\mathsfit}{bold}{\encodingdefault}{\sfdefault}{bx}{n}



%% file: preamble.tex
\definecolor{DarkGreen}{rgb}{0.55, 0.71, 0.0}
\definecolor{DarkRed}{rgb}{0.82, 0.1, 0.26}
\definecolor{darkblue}{rgb}{0, 0, 0.5}

\newcommand{\ours}{\text{ChartScope}\xspace} 
\newcommand{\oursb}{\text{ChartDQA}\xspace} 

%% file: sections/0_abstract.tex
Recent methods for customizing Large Vision Language Models (LVLMs) for domain-specific tasks have shown promising results in scientific chart comprehension. 
However, existing approaches face two major limitations: First, they rely on paired data from only a few chart types, limiting generalization to wide range of chart types. Secondly, they lack targeted pre-training for chart-data alignment, which hampers the model’s understanding of underlying data.
In this paper, we introduce \ours, an LVLM optimized for in-depth chart comprehension across diverse chart types. 
We propose an efficient data generation pipeline that synthesizes paired data for a wide range of chart types, along with a novel Dual-Path training strategy that enabling the model to succinctly capture essential data details while preserving robust reasoning capabilities by incorporating reasoning over the underlying data. Lastly, we establish \oursb, a new benchmark for evaluating not only question-answering at different levels but also underlying data understanding. Experimental results demonstrate that \ours significantly enhances comprehension on a wide range of chart types.\footnote{The code and data are available at the \href{https://davidhalladay.github.io/chartscope_demo}{project page}. \newline \hspace*{1.4em}{$^*$work done during research internship at Microsoft}}

%% file: sections/1_introduction.tex
\section{Introduction}

In today's data-driven world, visualizations like bar and pie charts play a crucial role in interpreting data. However, as data grows in volume and complexity, there is an increasing need for advanced tools that can improve our ability to process and analyze large-scale information efficiently.
Artificial Intelligence~(AI), particularly Large Vision Language Models~(LVLMs), is increasingly used to automate the understanding of scientific charts, promising more efficient and accurate analysis.
Robust benchmarks are also essential, setting standards and metrics that drive the development and evaluation of these AI tools.

Prior studies have introduced end-to-end neural models aimed at enhancing chart comprehension~\citep{lee2023pix2struct,liu2022matcha, zhou2023enhanced}, such as masked table prediction~\citep{zhou2023enhanced}, chart question answering~\citep{masry2023unichart}, and chart de-rendering~\citep{liu2022matcha}.
These models specialize in handling one task within the domain of chart analysis.
Furthermore, advancements in LVLMs, exemplified by LLaVA~\citep{liu2024visual, liu2023improved} and miniGPT~\citep{zhu2023minigpt}, have showcased versatility in vision-language tasks.
These generalist models undergo a two-stage training process: initially learning visual-language alignment through image-caption pairs, followed by end-to-end fine-tuning using image-QA pairs.
This training not only enables LLMs to interpret visual data but also retains their extensive pre-trained knowledge, which supports their reasoning abilities and leads to strong performance across diverse visual language tasks.

\input{tables/benchmark}

Recent advances have further ignited interest in tailoring LVLMs to specialized domains such as scientific chart understanding.
\citet{han2023chartllama,liu2023mmc} have explored collecting instruction-tuned chart data and low-rank adaptation~\citep{hu2021lora} to enhance LVLMs' proficiency with unique chart characteristics. However, due to scarcity of data of various chart types and its underlying data for fine-tuning, existing LVLMs struggles with not only understanding various chart types but also capturing underlying data when numerical values are not annotated.
We hypothesize that this issue stems from a gap in vision-language alignment between natural image-caption pairs and digital chart-data pairs.
Without targeted pre-training for chart-data alignment, models may resort to relying on a ``shortcut'' of recognizing numeric annotations through OCR, rather than truly understanding the visual subtleties of diverse charts.

To address the aforementioned challenges, in this paper we introduce \ours, a LVLM optimized for in-depth chart comprehension across many chart types. Specifically, we propose a novel data generation pipeline that leverages text-only LLMs to efficiently produce large-scale pairwise data covering various chart types, significantly reducing the cost and complexity of data generation for LVLM training. Secondly, by leveraging the synthesized data, we introduce a Dual-Path training strategy that enhances alignment between graphic and underlying data while preserving reasoning skills during fine-tuning. Combining the wide range of synthetic data with Dual-Path alignment training, \ours excels at interpreting various chart types (in-breadth) but also understanding the underlying data (in-depth).
Furthermore, existing chart benchmarks are limited in both chart and question types. This motivated us to introduce \oursb, a comprehensive chart benchmark comprising 20 types, 3 QA levels, and underlying data for each chart, designed to measure not only overall abilities but also the capability to capture underlying data.

%% file: tables/benchmark.tex
\begin{table*}[t]
\scriptsize
\centering
\vspace{-0.06in}
\newcolumntype{C}{>{\centering\arraybackslash}X}
\newcolumntype{L}{>{\raggedright\arraybackslash}X}
\newcolumntype{R}{>{\raggedleft\arraybackslash}X}
\begin{tabularx}{0.9\textwidth}{L r r r c c c}
\toprule
\multirow{2}{*}{Benchmark} & \multirow{2}{*}{\# Image} & \multirow{2}{*}{\# Chart type} & \multirow{2}{*}{\begin{tabular}[c]{@{}c@{}}Avg. \# QAs \\ per image\end{tabular}} & \multirow{2}{*}{\begin{tabular}[c]{@{}c@{}}Multi-level QAs \\ per image \end{tabular}} & \multirow{2}{*}{\begin{tabular}[c]{@{}c@{}}Raw data\\ per image\end{tabular}} & \multirow{2}{*}{\begin{tabular}[c]{@{}c@{}}Chart style\\ variation\end{tabular}} \\ 
 &  &  &  &  &  &  \\
\midrule
PlotQA~\citep{methani2020plotqa} & 33.7k & 3 & 1 & \xmark & \xmark & \xmark \\
ChartQA~\citep{masry2022chartqa} & 1.5k & 3 & 1 & \xmark & \cmark & \xmark \\
Chart-to-text~\citep{kantharaj2022chart} & 6.6k & 6 & 1 & \xmark & \xmark & \xmark \\
ChartBench~\citep{xu2023chartbench} & 2.1k & 9 & 9 & \cmark & \xmark & \xmark \\
ChartX~\citep{xia2024chartx} & 6k & 18 & 1 & \xmark & \cmark & \xmark \\
MMC~\citep{liu2023mmc} & 2k & * & 1 & \xmark & \cmark & \xmark \\
CharXiv~\citep{wang2024charxiv} & 2.3k & * & 5 & \cmark & \xmark & \xmark \\
EvoChart-QA~\citep{huang2025evochart} & 650 & 4 & 2 & \cmark & \xmark & \xmark \\
\cmidrule(lr){1-7}
Ours & 5.48k & 20 & 13.5 & \cmark & \cmark & \cmark \\ 
\bottomrule
\end{tabularx}
\vspace{-2mm}
\caption{Comparison with existing benchmarks for chart evaluation. *~denotes unbounded chart types. Chart variation denotes whether the dataset contains charts with different styles but sharing the same raw data.}
\label{table:benchmark_comparison}
\vspace{-6mm}
\end{table*}

%% file: sections/2_related_work.tex
\section{Related Works}

\noindent

\vspace{0.05in}
\noindent
Current approaches for LVLMs' chart understanding fall into two main categories: models specifically designed for chart-related tasks~\citep{lee2023pix2struct, zhou2023enhanced, masry2023unichart, liu2022matcha, masry2021integrating}, and those that utilize pre-trained LVLMs~\citep{masry2024chartinstruct, liu2023mmc, masry2024chartgemma, meng2024chartassisstant, chen2024onechart, zhang2024tinychart, xu2025chartmoe, huang2025evochart}.
The first group involves models trained exclusively on chart-specific data, often limited by the scope of the training datasets thus cannot be applied to diverse chart scenarios.
The second group, which involves adapting existing LLMs and LVLMs through fine-tuning~\citep{liu2024visual} or integration with external models~\citep{liu2022deplot}, shows promising versatility across various questions and scenarios.
Yet, research on developing methods for deep chart understanding across various types in practical settings remains scarce.
Additionally, models are evaluated against benchmarks focused on tasks like data extraction~\citep{masry2022chartqa, kantharaj2022opencqa, shi2024chartmimic}, summarization~\citep{kantharaj2022chart}, and basic mathematical reasoning~\citep{methani2020plotqa}, which predominantly feature basic chart types~(\eg, bar, line, pie charts) and lack nuanced differentiation in QA levels to thoroughly assess models’ understanding capabilities.
Recently, CharXiv~\citep{wang2024charxiv} and EvoChart~\citep{huang2025evochart} were introduced to evaluate general comprehension of real-world scientific charts. However, no existing benchmark targets the in-depth reasoning and understanding capabilities of multimodal LLMs.
Addressing these gaps, our work introduce a way to enhance in-depth and in-breadth chart understanding for LVLMs and a new benchmark with a variety of chart types, QA levels, and raw data to evaluate LVLMs’ comprehension abilities.

%% file: sections/3_method.tex
\vspace{-0.05in}
\section{In-Depth and In-Breadth Chart Understanding}

To build a chart understanding LVLM with in-depth and in-breadth understanding, a comprehensive dataset containing chart images paired with captions and raw data across various chart types is essential for pre-training and fine-tuning. However, no existing dataset provides the necessary variety of chart types, topics, and styles. To bridge this gap, we first introduce a novel data generation pipeline for large-scale chart data generation~(\cref{method:data}) and QAs generation~(\cref{method:qa}). With the synthetic data at hand, we can perform the feature alignment pre-training and end-to-end fine-tuning for LLMs.

\subsection{Quadratic-scale data generation}
\label{method:data} 

Our data generation leverages the promising text content generation and coding abilities of current large language models, \eg, GPT-4, to generate chart images and data. Specifically, LLMs allow us to synthesize raw data for charts, and then the generated Python script turns the raw data into a chart image. In this way, we can produce image data without accessing costly multimodal LLMs. Unlike previous works~\citep{han2023chartllama, xia2024chartx} that prompt LLMs to iteratively generate CSV data, QAs, and Python script for each chart image -- a process that is costly to massively scale -- our pipeline features parallel code and data generation through shared templates and READMEs for consistent definitions and formats across the same chart types. Most importantly, since all code script and data share the same structure, our generated data can be universally applied to any generated code and vice versa, significantly enhancing scalability without exhaustively prompting LLMs. We detail the pipeline further below.

\input{figures/framework}

\paragraph{Shared template and README.} As shown in~\cref{fig:framework}, given a chart type~(\eg, line) sampled from a predefined chart type database, the JSON expert LLM first generates a JSON template for the given chart type, along with a README file. In detail, the JSON template contains general information for the chart image, including the title, x-axis, y-axis information, and raw data. The README contains the definition of the chart type and the meanings of the keys and values to enhance understanding of the JSON template. Please refer to Sec.~\textcolor{darkblue}{G} for some examples. We note that the JSON template, together with the README, ensures the consistency of data generation so that further data and code generation can follow the explicit format and definition guidance of the template data. Note that we choose JSON as our primary data representation format, in contrast to previous works~\citep{han2023chartllama, masry2022chartqa, methani2020plotqa, xia2024chartx}, which used CSV. The JSON format allows us to incorporate not only numerical data but also additional chart information, such as titles and the scales of x and y axes, which is beneficial for pair-wise pre-training tasks. Moreover, JSON data is structured, and when paired with a README file, it minimizes ambiguity in data descriptions, which is particularly valuable for complex chart types. 

\paragraph{Orthogonal data and code generation.} With the template files at hand, we generate data and code independently. For the data generation branch, to ensure the generated data covers diverse topics, we jointly input the produced template files~(\ie, JSON template and README) and a topic sampled from a pre-defined topic set (\eg, energy production and market share) into a data expert LLM. For the complete topic list, please refer to Sec.~\textcolor{darkblue}{H}. We require the data expert LLM to follow the definitions in the template files and generate $M$ JSON data along with different kinds of questions and answers~(\eg, summary QA) based on the raw data. As for code generation, another code expert LLM is utilized to produce $N$ Python code based on the given chart type, data template, and Python library. Note that to prevent generating simple code repeatedly for the given chart type, we explicitly ask the code expert LLM to introduce visual variations in aspects such as color, legend, grid, font, and mark texture, \etc. More details can be found in Sec.~\textcolor{darkblue}{A}.

\paragraph{Diverse QA synthesis}
\label{method:qa} 

With the raw data for each chart as the input, we then use text-only LLM to generate question-answer~(QA) pairs for the instruction fine-tuning.
To cover various question-anwser for chart data, we include general QAs, containing not only description and summary QA but also three different level of QAs: literal QAs, inferential QAs, and reasoning QAs~(as illustrated in Fig.~\textcolor{darkblue}{A1}), encompassing a range of questions for chart images, covering abilities from basic data understanding and global concept comprehension to advanced reasoning. Please refer to the Sec.~\textcolor{darkblue}{A} for more details.


\vspace{-0.1in}
\paragraph{Composition for quadratically scaled data.}
As shown in Fig.~\textcolor{darkblue}{A2}, we consider 20 different chart types. For each chart type, we collect $N=400$ different Python codes and $M=1000$ different JSON data files covering various topics. Note that we perform automatic data filter based on predicted file structure's correctness, Python code execution errors, and OCR tools, refer to Table~\textcolor{darkblue}{A6} for more details. After filtering, we have $\approx5$ million images, with all the chart types listed in Fig.~\textcolor{darkblue}{A2}. For each chart image, we collect the raw data, a shared README, the corresponding Python script, $17$ general question-answer (QA) pairs: $1$ des. QA, $1$ summary QA, $5$ literal QAs, $5$ inferential QAs, $5$ reasoning QAs. Note that we use around 2 million synthetic data pairs to train the 13B model and 500k data pairs to train the 3B model. For the scaling law experiments, please refer to Sec.~\textcolor{darkblue}{F.2}.

\input{figures/framework_dp}

\subsection{Dual-Path training with augmented QAs}
With the aforementioned generated QAs, we can perform classical visual instruction tuning~\cite{liu2024visual}. However, unlike generic image understanding, chart image understanding requires the model to not only comprehend the underlying data of the chart but also perform reasoning to obtain the final answers. To enhance the in-depth understanding of the model, we introduce Dual-Path training (shown in~\cref{fig:framework_b}, which is built on top of the general chart QA pairs by including two additional augmented QAs (for training only): \textbf{Data-driven QAs} and \textbf{JSON-only QAs}. \textbf{Data-driven QAs} are multi-turn QAs that first prompt the model to extract JSON raw data given a chart and then answering the question based on the extracted JSON and chart. \textbf{JSON-only QAs} are instead a pure text QAs. Our goal is to preserve the reasoning ability of LLMs when extending to the chart domain. In practice, we replace images in the common QAs with JSON data and the README, so the models have to answer the questions based on the underlying data.
\input{tables/main}

\subsection{A new benchmark for comprehensive chart understanding}
\label{sec:benchmark}
A chart expert model should be capable of understanding a wide range of chart types and should not only be able to answer questions of varying complexity but also grasp the underlying data. However, as shown in~\cref{table:benchmark_comparison}, existing chart benchmarks either cover only a limited range of chart types~(\eg, line, bar, and pie charts) or lack comprehensive QA sets to evaluate a model's understanding of charts from various perspectives, including raw data comprehension, inferential abilities, and mathematical reasoning capabilities.
To bridge this gap, we propose \oursb, a benchmark derived from the aforementioned synthetic dataset.
It covers 20 different chart types, three different levels of QAs (literal, inferential, and reasoning QAs), and provides both long and short answers.
Notably, the chart images in the benchmark are not all annotated, allowing assessment of the model's ability to understand the underlying data of a chart as humans do.
To ensure the quality of the images in the benchmark, we employed human evaluations to filter the data and obtain a high-quality test set.
The evaluations are based on \textit{Answerability}
and \textit{Correctness}.
Please see Sec.~\textcolor{darkblue}{E} for more details about benchmark statistics, filtering, analysis, etc. Note that these QAs equally cover literal, inferential, and reasoning questions for measuring chart understanding of LVLMs.

%% file: figures/framework.tex

\begin{figure}[t]
  \centering
  \includegraphics[page=25, trim={160 100 160 100}, clip, width=0.45\textwidth]{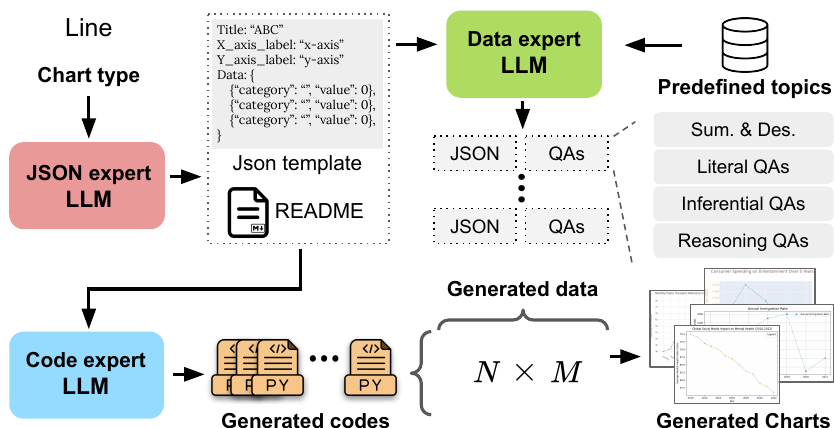}
  \vspace{-2mm}
  \caption{\textbf{Overview of the proposed data generation pipeline}.
  Generating code and data points conforming to a shared JSON template enables quadratic scaling of the data size~(\wrt to \#GPT calls).
  ($N$ and $M$ denote the number of generated scripts and data, respectively.)
  } 
  \label{fig:framework}
  \vspace{-6mm}
\end{figure}

%% file: figures/framework_dp.tex
\begin{figure}[t]
  \centering
  \includegraphics[page=26, trim={190 160 195 160}, clip, width=0.45\textwidth]{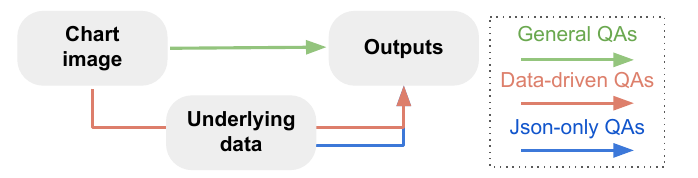}
  \vspace{-2mm}
  \caption{\textbf{Overview of the Dual-Path training strategies of \ours}.
  The Dual-Path training enforces the model to grasp the underlying data for chart question answering (via Data-driven QAs) while maintaining reasoning capability (via JSON-only QAs).
  } 
  \label{fig:framework_b}
  \vspace{-6mm}
\end{figure}

%% file: tables/main.tex
\begin{table*}[t]
\scriptsize
\centering
\vspace{-2mm}
\newcolumntype{C}[1]{>{\centering\arraybackslash}p{#1}}
\newcolumntype{L}[1]{>{\arraybackslash}p{#1}}
\begin{tabularx}{2.05\columnwidth}{p{3.95cm} c C{8mm} C{8mm} C{8mm} C{8mm} c c c c}
\toprule
\multirow{2}{*}{Method} & \multirow{2}{*}{\begin{tabular}[c]{@{}c@{}}\#\\Params\end{tabular}} & \multirow{2}{*}{MMC} & \multirow{2}{*}{ChartX} & \multicolumn{2}{c}{\oursb} & \multirow{2}{*}{PlotQA$^*$} & \multirow{2}{*}{ChartQA} & \multirow{2}{*}{\begin{tabular}[c]{@{}c@{}}Chart-\\to-Table\end{tabular}} & \multirow{2}{*}{\begin{tabular}[c]{@{}c@{}}Chart-\\to-Text\end{tabular}} \\
 \cmidrule(lr){5-6}
 &  &  &  & Basic & Adv. &  &  &  &  \\
\midrule
DePlot~{\citep{liu2022deplot}}              & -  & - & -    & -    & -    & -    & 79.3 & 87.2 & -    \\ 
ChartLlama~{\citep{han2023chartllama}} & 13B   & 0.55 & 13.8    & 23.5 & 18.0 & 29.8 & 69.7 & 89.8 & 14.2 \\
ChartInstruct{\citep{masry2024chartinstruct}} & 7B & 0.51 & 16.6 & 28.5 & 23.7 & 23.1 & 66.6 & 18.9 & 13.8 \\
ChartAst~{\citep{meng2024chartassisstant}} & 13B  & 0.57  & 31.0 & 28.6 & 22.7 & 26.2 & 79.9 & 91.6 & 15.5 \\
ChartGemma~{\citep{masry2024chartgemma}} & 3B  & 0.57  & 17.2 & 11.2 & 9.8 & 6.2 & 80.2 & - & - \\
ChartMoE@490$^{*}$~{\citep{xu2025chartmoe}} & 8B  & \textbf{0.77} & 30.6 & 34.2 & 28.5 & 17.1 & 81.2 & - & - \\
TinyChart@768~{\citep{zhang2024tinychart}} & 3.1B  & 0.57 & \underline{33.4} & 27.7 & 22.1 & 32.6 & \textbf{83.6} & \textbf{93.8} & \underline{17.2} \\
\midrule
\ours$_{\text{LLaVA-7B}}$ & 7B & 0.52 & 27.6 & 42.2 & 33.3 & 30.1 & 70.0 & 83.6 & 11.5 \\
\ours$_{\text{LLaVA-13B}}$ & 13B & 0.54 & 31.4 & \underline{45.1} & \underline{37.3} & \underline{34.0} & 71.4 & 88.1 & 12.7 \\
\ours$_{\text{TinyLLaVA-3.1B@768}}$ & 3.1B  & \underline{0.59} & \textbf{35.7} & \textbf{47.1} & \textbf{38.3} & \textbf{35.2}  & \underline{83.2} & \underline{93.3} & \textbf{17.4} \\
\bottomrule
\end{tabularx}
\vspace{-2mm}
\caption{\textbf{Comprehensive evaluation across various chart benchmarks.}
\ours achieves best QA results on both (mostly) advanced benchmarks (i.e., MMC, ChartX, and \oursb) and non-annotated benchmark, PlotQA.
Basic chart types in \oursb denotes bar, line, and pie charts.
$^{*}$ denotes MMC training set are used in the model training.
The best result is highlighted in \textbf{Bold} and the second \underline{underlined}. }
\vspace{-5mm}
\label{exp:main} 
\end{table*}

%% file: sections/4_exp.tex
\section{Experiments and Model Analysis}

\subsection{Experimental setup}
\paragraph{Benchmark.}
We test our model on seven chart benchmarks and compare it against previous works. These include recent benchmarks with advanced chart types, such as MMC
(VQA split~\cite{liu2023mmc}), ChartX
(VQA track~\cite{xia2024chartx}), and~\oursb, classical benchmarks with annotated charts such as PlotQA~\citep{methani2020plotqa}, and non-annotated charts such as ChartQA~\citep{masry2022chartqa}. For benchmark details and evaluation metrics, we follow each benchmark’s protocol; please refer to Sec.~\textcolor{darkblue}{B} for more information. For additional comparison results on MMC, EvoChart~\cite{huang2025evochart}, and ChartBench~\cite{xu2023chartbench}, please see Sec.~\textcolor{darkblue}{F}.

\vspace{-0.1in}
\paragraph{The details of training process.}
We train all models in three stages: First, we pretrain the projector and then jointly fine-tune the model end-to-end following the classical LLaVA approach~\cite{liu2024visual}. Finally, we perform chart-specific downstream (LoRA) fine-tuning.
Specifically, in the initial pretraining stage, we train only the projector using the original LLaVA data alongside our newly generated chart descriptions and chart-JSON pairs. Next, we fine-tune both the projector and the LLM using the original LLaVA QA pairs together with our generated chart QA pairs.
Finally, we apply downstream fine-tuning to align the LLM’s response distribution with that of the target chart dataset. For the LLaVA version of \ours, due to computational constraints, we perform LoRA fine-tuning on each benchmark separately. For TinyLLaVA~\cite{zhou2024tinyllava}, we perform standard fine-tuning using the TinyChart dataset~\cite{zhang2024tinychart} for a fair comparison.
Please refer to Table \textcolor{blue}{1} in the TinyChart paper for more details. Each stage is carefully studied, and the results are presented in the following subsections.



\subsection{Main comparison}

We compare \ours with previous chart domain specific models as the results shown in~\cref{exp:main}. For comparison of non chart expert models, please refer to Table~\textcolor{darkblue}{A6}.

\paragraph{Question-answering on various chart types.}
We first evaluate performance on MMC and ChartX to showcase our model's ability to understand a wide range of chart types. The MMC benchmark contains real chart data collected from academic articles with unbounded chart types, while ChartX contains synthetic data with 18 chart types. As shown in~\cref{exp:main}, our model achieves the second-best performance on MMC--behind ChartMoE, explicitly fine-tuned with MMC’s training data--and outperforms previous works on ChartX by approximately $2\%$. Additionally, we report results on \oursb for both basic and advanced chart types. Our performance on advanced types consistently outperforms previous works, verifying the effectiveness of our approach. For underlying data evaluation and comparison on~\oursb, please refer to Sec.~\textcolor{darkblue}{E}.

\paragraph{Performance on unannotated chart images.}
Most of the images in ChartQA~\citep{masry2022chartqa} are annotated, which means the numerical values of data points are explicitly shown on the images.
However, real-world charts may be unannotated, requiring models to capture the underlying data rather than relying solely on OCR. To measure chart understanding in these scenarios, we further evaluate models using the PlotQA dataset, and the results are shown in~\cref{exp:main}. Notably, since training previous models like ChartLlama on PlotQA is infeasible, we load the model weights used in ChartQA and perform zero-shot prediction on PlotQA.
The results show that our model performs significantly better~($\approx+3\%$) on unannotated chart images than the previous SOTA, TinyChart, suggesting that our training methods rely less on numerical annotations.

\paragraph{Performance on classical benchmarks.}
We now compare performance on classical benchmarks, such as ChartQA, Chart-to-Table, and Chart-to-Text. As shown in~\cref{exp:main}, \ours achieves on-par accuracy with the SOTA on ChartQA. Additionally, \ours achieves a competitive F1 score on Chart-to-Table, indicating that it can capture not only the structure but also the numerical values of raw chart data. We note that performance on these benchmarks may be saturated, as the images are mostly annotated and chart types are limited. In this context, these benchmarks primarily measure OCR capability and do not assess the ability to capture the underlying data. As for Chart-to-Text, as shown in~\cref{exp:main}, \ours performs comparably in capturing global concepts and can caption chart images with meaningful text. For qualitative examples, please see Sec.~\textcolor{darkblue}{F.7}.



\input{tables/ablation_pretrain}
\input{tables/ablation_finetune}
\subsection{Ablation study}
\subsubsection{Chart feature alignment pre-training}
\label{method:pretraining}
To study the effectiveness of pretraining using generated pair-wised data, we compare three configurations: utilizing only LLaVA CC3M Pretraining data,
combining LLaVA data with chart-description pairs, and using LLaVA data with both chart-description and chart-raw data pairs. The data for stage two training remains consistent across these settings, summary QAs, description QAs, three-level QAs, text-only QAs, and data-driven QAs. 
We use LLaVA-7B as the baseline for this comparison, and the results are detailed in \cref{exp:abltion_pretrain}. We found that dense data alignment is beneficial for both chart data comprehension and reasoning. Specifically, utilizing chart-json pairs in the pre-training of projector improve the human split of ChartQA by 4\% on top of the performance of using classical chart-caption pairs.


\subsubsection{Dual-Path fine-tuning}
\label{method:finetuning}
We investigate the effectiveness of the data used in end-to-end fine-tuning, including the introduced Dual-Path training data. We conduct ablation studies starting with a baseline that uses only LLaVA Instruct-150K data,
incrementally adding extra QA pairs, and the results are shown in~\Cref{exp:abltion_finetune}.
Note that all methods leverage the same pre-training weights, derived from training on LLaVA data with both chart-description and chart-raw data pairs~(the best setting in~\cref{method:pretraining}). 
Our assumption for JSON-QAs is that, with a well-aligned first stage of training, re-blending some pure textual QAs can preserve the ability of reasoning on text raw data and also benefit the reasoning abilities in visual-text scenarios. As shown in~\cref{exp:abltion_finetune}, we discovered that re-blending JSON-only data during the end-to-end fine-tuning stage improves chart reasoning skills by $3\%$ on the human split of ChartQA. Additionally, we study the effectiveness of Data-driven QAs, which are multi-turn QAs requiring models to extract raw data before answering questions. We find that, combined with the raw data reasoning abilities enhanced via JSON-only QAs, models achieve better reasoning robustness and overall performance, verifying the effectiveness of our design. Furthermore, leveraging data prompting in inference, requiring model extract raw data and then answering the question, significantly improves performance across all downstream tasks.

%% file: tables/ablation_pretrain.tex
\begin{table}[t]
\footnotesize
\centering
\vspace{-2mm}
\newcolumntype{C}{>{\centering\arraybackslash}X}
\newcolumntype{L}{>{\raggedright\arraybackslash}X}
\begin{tabularx}{1.0\columnwidth}{L cc}
\toprule
\multirow{2}{*}{Training data}  & \multicolumn{2}{c}{ChartQA}  \\  
\cmidrule(lr){2-3} 
 & human & augmented \\
\midrule
LLaVA-CC3M-Pretrain pairs & 44.80 & 83.92 \\
\hspace{3mm}+ Chart-description pairs & 48.56 & 86.89 \\
\hspace{6mm}+ Chart-JSON data pairs & \textbf{52.28} & \textbf{87.68} \\
\bottomrule
\end{tabularx}
\vspace{-2mm}
\caption{\textbf{Ablation of pretraining data.}
This empirically verifies that pre-training basic chart visual perception is still important, even with abundant stage-2 instruction fine-tuning data. Moreover, learning to predict JSON data is beneficial even on top of pre-training with descriptive captions.}
\vspace{-3mm}
\label{exp:abltion_pretrain}
\end{table}


%% file: tables/ablation_finetune.tex
\begin{table}[t]
\footnotesize
\centering
\vspace{0mm}
\newcolumntype{C}{>{\centering\arraybackslash}X}
\newcolumntype{L}{>{\raggedright\arraybackslash}X}
\begin{tabularx}{1.0\columnwidth}{L cc}
\toprule
\multirow{2}{*}{Training data}  & \multicolumn{2}{c}{ChartQA}  \\  
\cmidrule(lr){2-3} 
 & human & augmented \\
\midrule
LLaVA-Instruct-150K QAs & 45.84 & 86.48 \\
\hspace{6mm}+ General QAs & 48.96 & 87.52 \\
\hspace{9mm}+ JSON-only QAs & 49.60 & 87.36 \\
\hspace{12mm}+ Data-driven QAs & \underline{52.28} & \textbf{87.68} \\
\cmidrule(lr){1-3}
\hspace{15mm}+ Data Prompting$^\dagger$ & \textbf{56.96} & \underline{87.60} \\
\bottomrule
\end{tabularx}
\vspace{-2mm}
\caption{\textbf{Ablation of Dual-Path training.}
Each type of new instruction/QA data improves the final performance consistently across almost all metrics. Best result is highlighted in \textbf{Bold} and the second best is \underline{underlined}. $^\dagger$ denotes an inference technique without extra data. General QAs contains description, summary, literal, inferential, and reasoning QAs.}
\vspace{-6mm}
\label{exp:abltion_finetune}
\end{table}

%% file: sections/5_conclusion.tex
\section{Conclusion}
In this paper, we introduce \ours, a Multimodal Large Language Model (LVLM) tailored for in-depth and in-breadth chart understanding. Powered by a data generation pipeline and a Dual-Path training strategy, our model is capable of interpreting diverse chart types independently of numerical annotations. Extensive experiments confirm that \ours surpasses the previous state-of-the-art across multiple benchmarks, validating the effectiveness of our framework. Additionally, we present a new benchmark specifically designed to evaluate LVLMs' comprehension across various chart types and multiple levels of understanding.
\pagebreak
\section{Limitations and Social Impact}
In this paper, we propose an LVLM model for chart understanding, fundamentally trained on synthetic data. However, since the synthetic data generated by LLMs cannot be perfect, sometimes incorrect data can be introduced into the dataset and may not be filtered out by our filtering process. These data can result in misalignments and incorrect mappings during pre-training and fine-tuning, potentially leading to incorrect responses and hallucinations. Thus, the performance of our chart LVLM is limited by the LLMs' generation capabilities. We can potentially include more advanced LLMs in the data generation pipeline to reduce the occurrence of incorrect data. Moreover, another limitation of our model is that it currently supports understanding only 18 chart types. However, there are many more chart types in the real world. Developing an open-domain, versatile chart understanding LVLM remains a task for future work.

\paragraph{Social impact}
Our model is capable of chart understanding and can interpret the raw data of a chart like a human, without relying on annotations, while also performing various levels of QA tasks. Thus, our model can be used in many data analysis scenarios, such as market research, healthcare trend analysis, and other data science areas. With the help of our model, humans can process large volumes of chart data more efficiently, make informed decisions, and enhance reporting accuracy. While our model provides benefits in chart understanding and analysis, there are potential negative impacts. For instance, it could be employed to create misleading data visualizations or generate false narratives when combined with other LLM tools. These fake charts and pieces of information can negatively affect decision-making processes.
 